\documentclass[10pt,twocolumn,letterpaper]{article}

\usepackage[normalem]{ulem}
\usepackage{wacv}              

\usepackage{amsmath}
\usepackage{graphicx}
\usepackage{pifont}
\newcommand{\cmark}{\ding{51}}
\newcommand{\xmark}{\ding{55}}

\usepackage[ruled,vlined]{algorithm2e}
\usepackage{float}
\usepackage{xcolor}
\definecolor{commentcolor}{RGB}{156,156,156}
\newcommand{\comment}[1]{\textcolor{commentcolor}{\# #1}}

\definecolor{wacvblue}{rgb}{0.21,0.49,0.74}
\usepackage{xcolor}

\definecolor{red}{HTML}{DB4F59}
\usepackage[pagebackref,breaklinks,colorlinks,allcolors=wacvblue]{hyperref}

\def\wacvPaperID{587} 
\def\confName{WACV}
\def\confYear{2026}

\title{Exploiting Label-Independent Regularization from Spatial Patterns for Whole Slide Image Analysis}

\author{
Weiyi Wu$^{1}$, Xinwen Xu$^{2}$, Chongyang Gao$^{3}$, Xingjian Diao$^{1}$, \\
Siting Li$^{1}$, and Jiang Gui$^{1}$\\
$^{1}$Dartmouth College; $^{2}$Massachusetts General Hospital; $^{3}$Northwestern University\\
{\tt\small \{weiyi.wu.gr, xingjian.diao.gr, siting.li, jiang.gui\}@dartmouth.edu} \\ 
{\tt\small xixu3@mgh.harvard.edu,}  \\
{\tt\small chongyanggao2026@u.northwestern.edu}
}

\begin{document}
\maketitle
\begin{abstract}
    Whole slide images, with their gigapixel-scale panoramas of tissue samples, are pivotal for precise disease diagnosis. However, their analysis is hindered by immense data size and scarce annotations. Existing MIL methods face challenges due to the fundamental imbalance where a single bag-level label must guide the learning of numerous patch-level features. This sparse supervision makes it difficult to reliably identify discriminative patches during training, leading to unstable optimization and suboptimal solutions. We propose a spatially regularized MIL framework that leverages inherent spatial relationships among patch features as label-independent regularization signals. Our approach learns a shared representation space by jointly optimizing feature-induced spatial reconstruction and label-guided classification objectives, enforcing consistency between intrinsic structural patterns and supervisory signals. Experimental results on multiple public datasets demonstrate significant improvements over state-of-the-art methods, offering a promising direction. The code is available at https://github.com/wwyi1828/SRMIL.
\end{abstract}    
\section{Introduction}
\label{sec:intro}

\begin{figure}[t]
\centering
\includegraphics[width=0.5\textwidth]{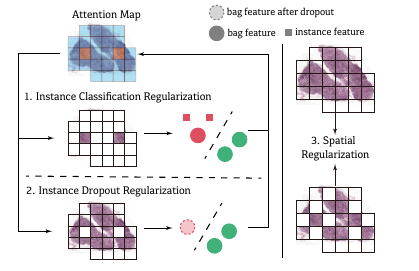}
\caption{Illustration of regularization fashions in MIL. In attention maps, orange represents high-attention instances. In the classification diagrams, red and green represent the target and non-target classes, respectively. The dense dashed lines represent the decision boundary. Circles indicate bag features, squares denote instance features, and dashed circles represent bag features after dropout. (1) Enforces label consistency between high-attention instances and bags. (2) Stochastically drops instances based on attention scores. Both methods are label-driven and iteratively update the model based on classification, with the updated model generating new attention maps for the next iteration. (3) Regularizes feature spaces through masked feature reconstruction, offering a label-independent and noise-free regularization.}
\label{fig:regularization_techniques}
\vspace{-10pt}
\end{figure}

The advent of Whole Slide Images represents a transformative advancement in medical diagnostics, offering pathologists unprecedented capabilities to examine tissue samples with remarkable detail and precision. Unlike conventional images, WSIs are characterized by their gigapixel resolution, typically encompassing $100,000 \times 100,000$ pixels at the 40x magnification level, enabling comprehensive tissue analysis critical for accurate disease diagnosis and prognosis. However, this level of detail introduces significant challenges for both manual and computational analysis.

In clinical practice, the vast scale of WSIs presents two fundamental challenges. First, acquiring precise pixel-level annotations demands substantial expertise and time investment from specialized pathologists ~\cite{abels2019computational, chen2021annotation}. Second, even when such detailed annotations are available, training on WSIs demands substantial computational resources and involves intricate training strategies~\cite{liu2017detecting,wang2016deep}. A few hundred WSIs contain as many pixels as the entire ImageNet dataset. These challenges are further compounded by the long-tailed distribution of regions within WSIs - when analyzed at the patch level, diagnostically relevant regions could constitute only a small fraction while the majority consists of normal or background tissue.

\begin{figure*}[t]
  \centering
  \includegraphics[width=1\textwidth]{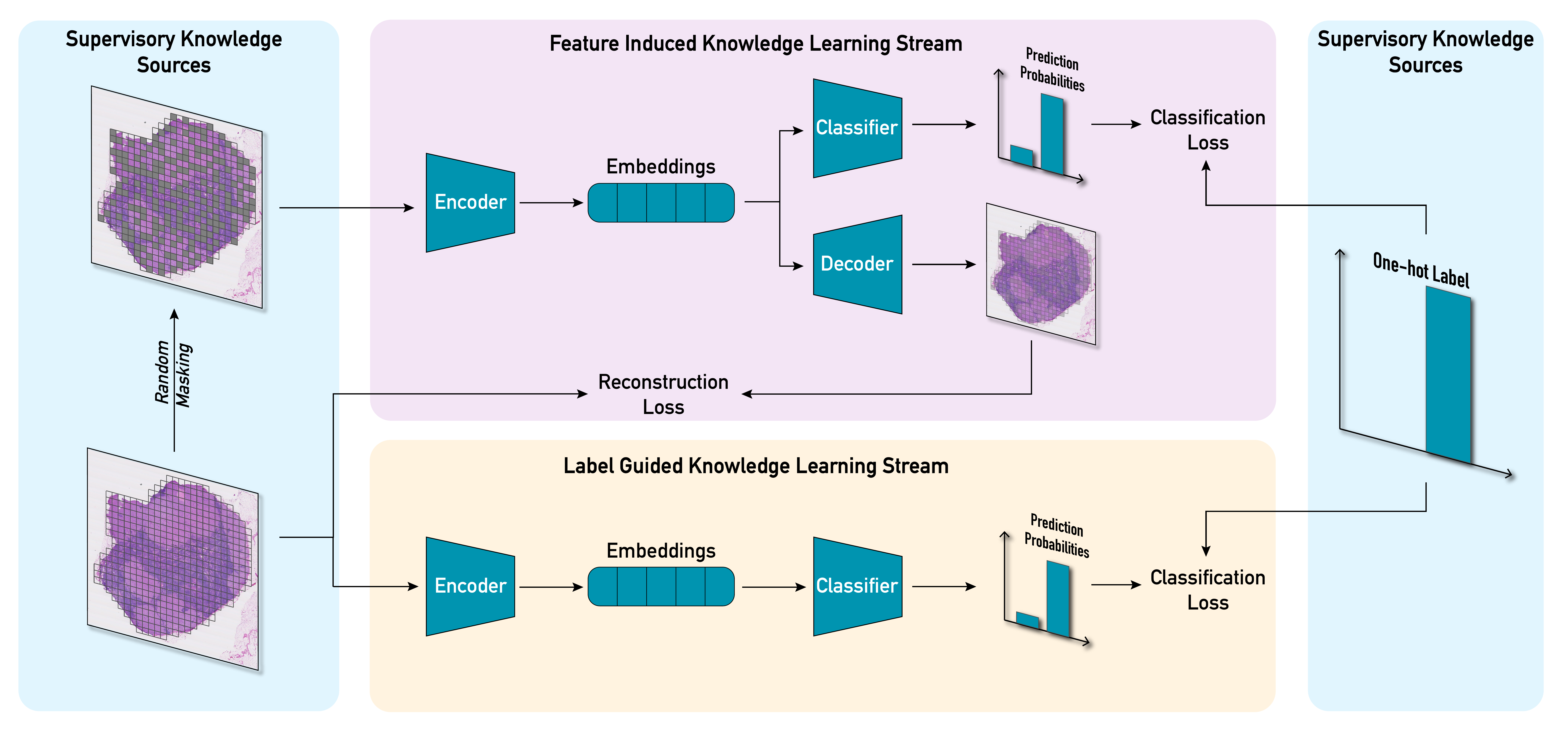}
  \caption{Overview of the SRMIL framework. The model employs dual learning streams: a feature-induced stream with self-supervised reconstruction and a label-guided classification stream. The encoder-decoder architecture processes WSI patch embeddings through graph attention networks, where random masked patch features reconstruction serves as noise-free regularization, complementing the supervised learning objective. Black lines indicate information flow from input patches through both learning streams.}
  \label{fig:pipeline}
\end{figure*}

To address these, the field has embraced weakly supervised approaches, particularly multiple instance learning (MIL)~\cite{wang2023dual, campanella2019clinical, ilse2018attention}. MIL methods operate by extracting features from fixed-size patches and learning from slide-level annotations, significantly reducing both annotation burden and training complexity. This paradigm has made WSI analysis more tractable and scalable, enabling models to predict slide-level outcomes by aggregating information across multiple patches under slide-level supervision.

However, these weakly supervised MIL approaches face their own set of limitations. The fundamental challenge lies in the inherent nature of WSI datasets: they typically contain only a few hundred samples, most lacking pixel-level annotations, yet each slide generates tens of thousands of high-dimensional patch features. This combination of limited supervision and flexible feature spaces creates a risk: models can easily learn spurious patterns specific to the training set rather than genuinely discriminative features. This often leads to overfitting and poor generalization of unseen data. To combat these issues, researchers have proposed various regularization strategies through auxiliary loss terms~\cite{Tang_2023_ICCV,lu2021data,zhang2022dtfd,zhang2023attention}, which can be viewed as regularization terms or techniques to constrain the overly flexible feature space (Fig.~\ref{fig:regularization_techniques}). However, existing regularization methods often rely on potentially noisy or incorrect supervisory signals, limiting their effectiveness. Furthermore, while these methods attempt to extract additional supervision signals, they remain fundamentally confined to classifier-driven constraints, overlooking the rich label-independent correlations between patch coordinates and features inherent in WSIs.

Recent advances in self-supervised learning (SSL) offer a promising direction for more noise-free regularization objectives. SSL has demonstrated remarkable success in learning well-structured and meaningful representations without relying on the supervision from manual annotations, including natural language processing \cite{devlin2018bert, brown2020language} and computer vision \cite{chen2020simple, dwibedi2021little, he2022masked, he2020momentum, he2019rethinking}. Through carefully designed pretext tasks, SSL enables models to learn rich representations by exploiting inherent data structure, which is independent of labels. Moreover, research has shown that combining SSL with supervised learning leads to more robust and generalizable representations~\cite{tan2022hyperspherical}, making it a promising solution to address both label sparsity and the curse of dimensionality in WSI analysis.

Motivated by these observations, we present a novel spatially regularized framework that leverages both supervised and self-supervised learning for WSI analysis Fig.~\ref{fig:pipeline}). Our work makes the following contributions:
\begin{enumerate}
    \item We introduce a dual-path learning architecture that integrates Graph Attention Networks (GAT)~\cite{veličković2018graph} with self-supervised reconstruction, leveraging inherent structural information in WSIs through a label-independent regularization mechanism.

    \item We demonstrate that self-supervised signals can serve as effective regularization mechanisms in weakly supervised scenarios, providing a new paradigm for leveraging abundant unlabeled data in medical image analysis.

    \item Through comprehensive experiments on multiple WSI classification tasks, we empirically validate that our approach significantly outperforms existing methods, providing evidence that integrating spatial information with self-supervised learning can substantially improve both accuracy and generalization in computational pathology.
\end{enumerate}

\section{Related Works}
\subsection{Self-Supervised Learning}

    Self-supervised learning (SSL) methods have gained prominence in various domains, aiming to learn meaningful representation spaces without relying on labeled data. It has two main fashions. Contrastive learning encourages models to learn invariant feature representations across diverse views while maintaining discriminative power between distinct samples. This approach optimizes for both uniformity and alignment in the representation space, with successful applications in visual~\cite{chen2020simple, he2020momentum, wu2023improving}, textual~\cite{gao2021simcse, jian2022non}, and graph data~\cite{you2020graph, veličković2018deep}. Generative paradigms, such as autoencoders~\cite{vincent2008extracting, kingma2013auto} and bidirectional GANs~\cite{donahue2017adversarial, donahue2019large}, focus on reconstructing data from latent spaces, learning meaningful representation spaces in the process. The landscape of SSL underwent a profound transformation with the introduction of transformer architectures and novel pretext tasks such as Masked Language Modeling and autoregressive language modeling in NLP~\cite{devlin2018bert, brown2020language}. This inspired Masked Image Modeling in computer vision, where models like Masked Autoencoders~\cite{he2022masked} mask random patches and reconstruct them, developing a deep understanding of spatial and contextual relationships. Similarly, in graph learning, Similarly, masked modeling objectives have been adopted in graph-structured learning and extended to incomplete supervision settings, where masked or partial-view prediction captures intrinsic dependencies and promotes representation consistency~\cite{li2023s, hou2022graphmae,xie2024uncertainty,xie2025multi}. These advancements highlight the importance of SSL in learning transferable and expressive feature representations across domains and suggest its potential as a principled regularization mechanism.

    \subsection{Multiple-Instance Learning} 

    Recent advances in computational pathology have established Multiple Instance Learning (MIL) as a dominant paradigm for analyzing WSIs, where individual patches are encoded into feature vectors and aggregated to create bag-level features. Attention-based multiple-instance learning (ABMIL)~\cite{ilse2018attention} marks a milestone by introducing a learnable attention mechanism that automatically identifies and weights informative instances during the end-to-end training process. However, ABMIL and other existing MIL methodologies face inherent challenges in analyzing WSIs, particularly overfitting due to limited label information.
    
    Various extensions of ABMIL have been proposed to address these challenges. Some methods focus on incorporating additional information flows, such as adjusting MIL training with bag contextual priors~\cite{lin2023interventional} or combining key instance identification with attention-based pooling~\cite{li2021dual}. Others introduce regularization strategies to improve model robustness. A common approach is to mask out instances based on their attention scores to prevent the model from overfitting to a few key instances~\cite{Tang_2023_ICCV, zhang2023attention, qu2022bi}. Some methods enforce label consistency by ensuring highly-attended instances share the same label as their bags~\cite{lu2021data}, while others decompose WSIs into sub-bags with shared labels for additional supervision~\cite{zhang2022dtfd}. However, these attention-based regularization approaches rely on the quality of attention weights and label assumptions, which may not always be reliable in practice. This is particularly challenging in histopathological image analysis where regions of interest could occupy only a small fraction of the entire dataset. Consequently, the effectiveness of these methods often depends on careful hyperparameter tuning and prior knowledge about the proportion of informative regions, risking the introduction of inaccurate supervisory signals when such proportions vary across different datasets. DGRMIL~\cite{zhu2024dgr} has also highlighted the importance of regularization by introducing an attention-independent regularization. However, the information flow for supervision remains label-exclusive, without exploiting the rich inherent signals to constrain the latent space.

    An alternative approach focuses on leveraging structural and positional relationships among instances. Graph Neural Networks (GNNs) transform the MIL problem into one where spatial relationships between instances are explicitly modeled~\cite{chen2021whole, wu2023graph}, in which each patch is represented as a node and the entire slide constitutes a graph. Transformer architectures have also been investigated for modeling instance interactions, though often limited by computational complexity. The Graphformer~\cite{zheng2022graph}, while innovative, is challenged by high memory demands due to its use of dense adjacency matrices and transformer layers, limiting its application to large WSIs.  While TransMIL~\cite{shao2021transmil} and MambaMIL~\cite{yang2024mambamil} leverage advanced architectures such as the Nystromformer~\cite{xiong2021nystromformer} and Mamba~\cite{gu2023mamba, zheng2025gmmamba}, capturing spatial relationships while maintaining manageable efficiency. Notably, while these approaches demonstrate the value of incorporating spatial information in the training process, they have not explored its potential as an explicit regularization mechanism. This leaves an important opportunity unexplored: leveraging the inherent structural patterns in WSIs as a reliable source of regularization for model training.

\begin{figure*}
    \centering
    \begin{subfigure}{.236\textwidth}
        \centering
        \includegraphics[width=\linewidth]{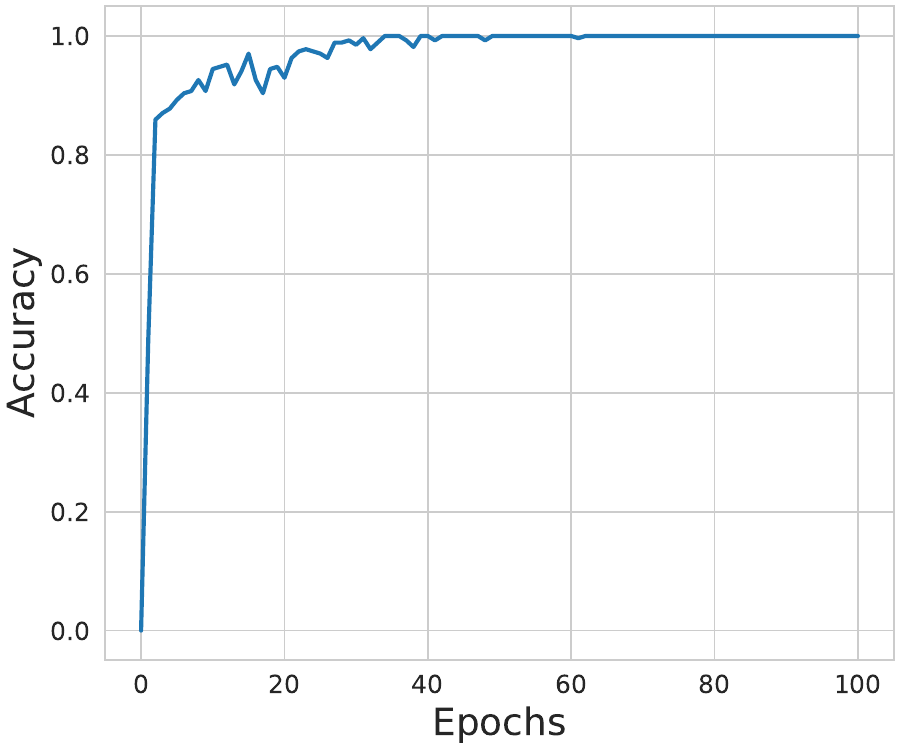}
        \caption{}
        \label{fig:train_accs}
    \end{subfigure}
    \begin{subfigure}{.236\textwidth}
        \centering
        \includegraphics[width=\linewidth]{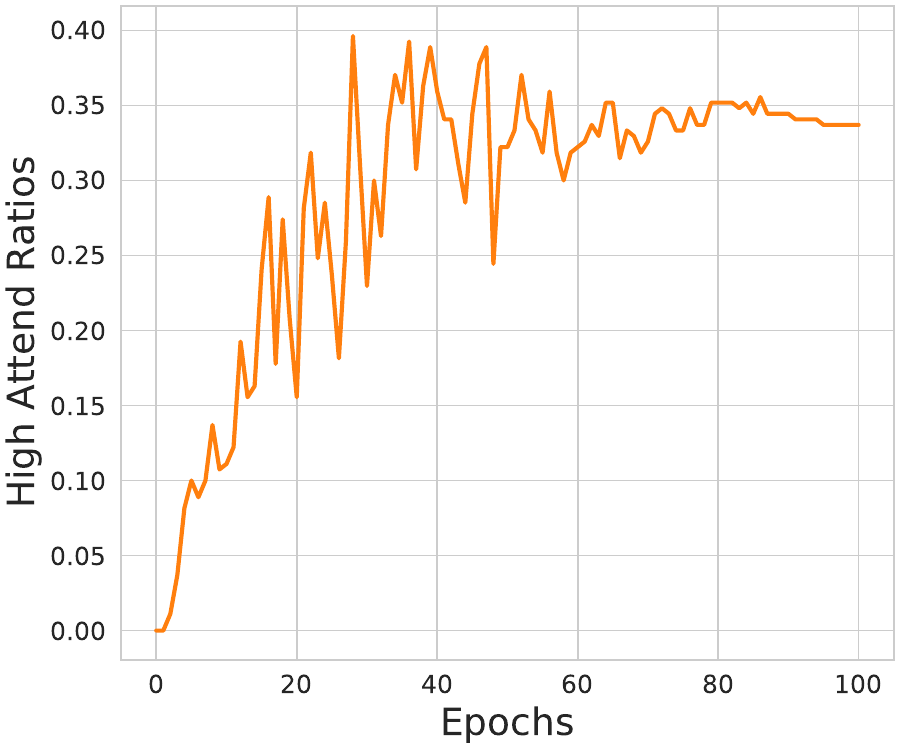}
        \caption{}
        \label{fig:attend_ratios}
    \end{subfigure}
    \begin{subfigure}{.236\textwidth}
        \centering
        \includegraphics[width=\linewidth]{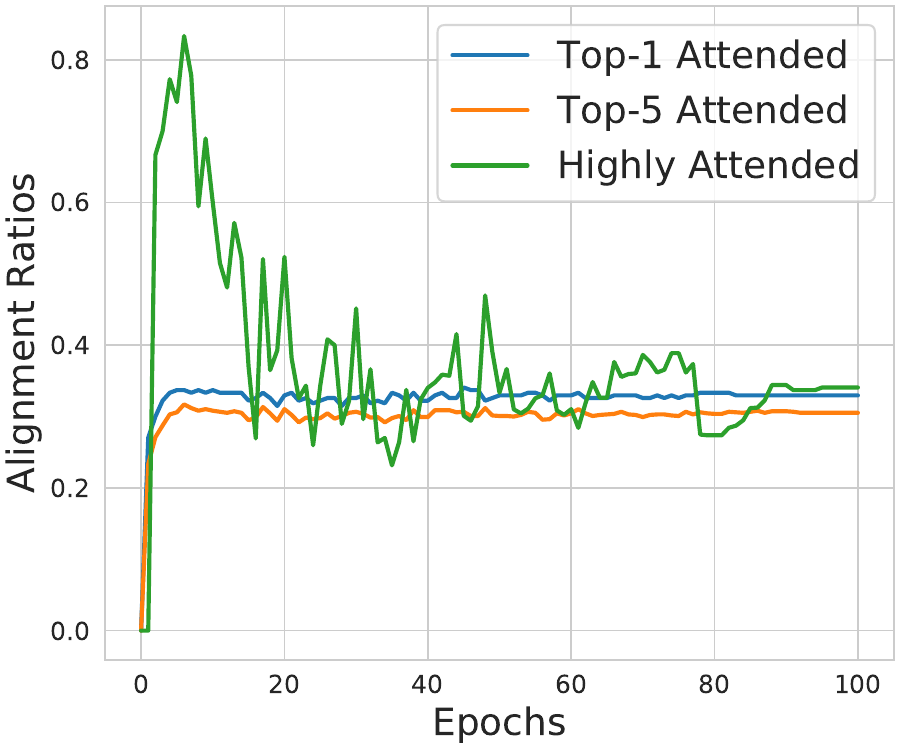}
        \caption{}
        \label{fig:alignment_ratios}
    \end{subfigure}
    \begin{subfigure}{.236\textwidth}
        \centering
        \includegraphics[width=\linewidth]{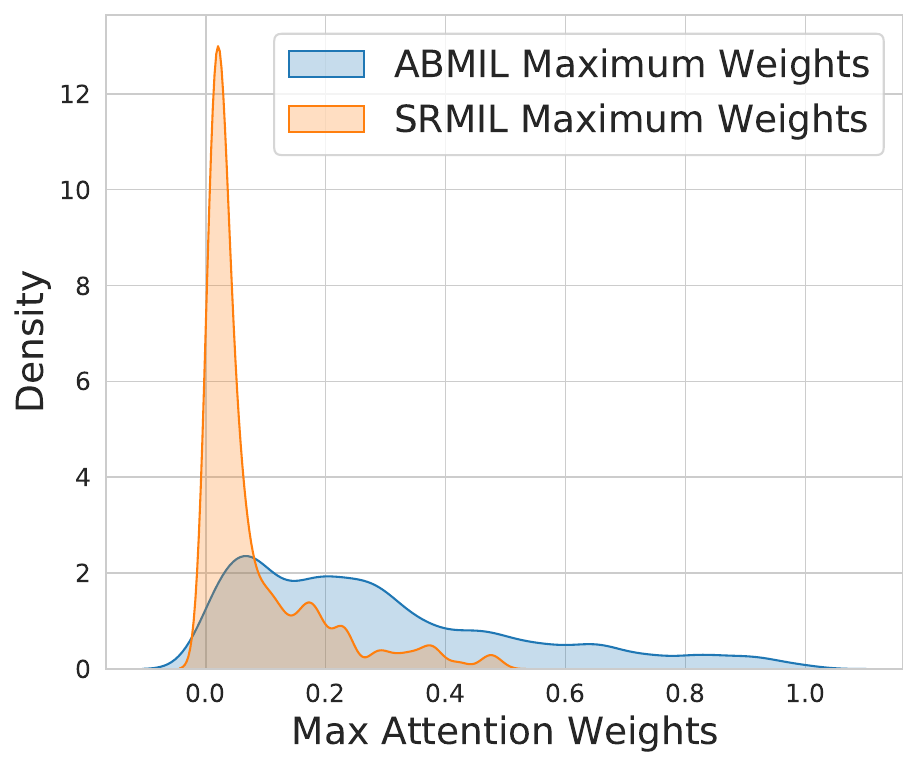}
        \caption{}
        \label{fig:attn_dist}
    \end{subfigure}

        \caption{
        Behavior Analysis of the ABMIL Model on the training set during training.
        \textbf{(a)} Training accuracy curve. 
        \textbf{(b)} Ratio of WSIs containing highly attended instances (attention weights $\geq$ 0.5). 
        \textbf{(c)} Instance-bag label alignment ratios for positive slides, categorizing instances as top-1 attended, top-5 attended, and highly attended. An instance is labeled positive if over 20\% of its area contains annotated positive regions. Negative slides are excluded as they contain only negative instances.
        \textbf{(d)} Distribution of maximum attention weights. ABMIL exhibits a highly skewed attention pattern with maximum weights up to 1, whereas SRMIL's weights are concentrated below 0.1, indicating a more uniform attention distribution.
    }

    \label{fig:training_metrics}
\end{figure*}

\section{Method}

    \subsection{Motivation}
    Given the limited number of labeled WSIs compared to their rich feature space, a key challenge lies in effectively regularizing the learning process. Recent efforts introduce additional regularization objectives to the ABMIL framework, typically relying on attention scores learned through slide-level supervision and predefined hyperparameters to guide either instance filtering or bag-level consistency constraints. However, such label-driven techniques could introduce noise in various scenarios: when slides contain few positive instances, the number of instances selected by the top-k approach may exceed the actual positives, leading to negative instances being incorrectly treated as positive during training. Conversely, all positive instances may be discarded, resulting in positive bags being trained with only negative instances. Both situations introduce erroneous supervision signals that hinder effective regularization, particularly challenging in datasets where regions of interest often occupy only a small fraction of the entire dataset.
    
    To validate these concerns, we analyze the behavior of ABMIL during training on the CAMELYON-16 dataset. Our analysis reveals several concerning patterns: While the model quickly achieves perfect training accuracy, two concerning patterns emerge. The ratio of highly attended instances continues to increase, yet the alignment ratio, which measures how often the highly attended instances are truly positive according to ground-truth instance labels, remains consistently low, suggesting the model may be memorizing negative instances as positive to overfit the dataset. The skewed attention distribution further demonstrates how the attention mechanism tends to focus on a small subset of instances overly. Notably, despite these potential issues, label-driven regularization methods still improve model generalization on unseen data, indicating that additional regularization is helpful even if noisy. This observation leads us to hypothesize that clean, label-independent regularization should lead to more robust representations without the risk of introducing noise.
    
    Motivated by this hypothesis, we propose to leverage the inherent spatial relationships among patch features as a knowledge source for regularization. We present our Spatially Regularized Multiple-Instance Learning (SRMIL) methodology, which consists of two complementary learning streams: a label-guided stream for supervision and a feature-induced stream that exploits spatial relationships as label-independent regularization. This dual-stream design enhances patch representations and reduces the feasible parameter space by learning from both label supervision and intrinsic spatial patterns.
    
    The proposed approach offers two key advantages over existing methods. First, unlike previous regularizations that rely on potentially noisy label-driven attention scores, our feature-induced stream provides additional supervision that is inherently noise-free and reliable, helping the model learn structural representations by leveraging the natural organization of WSIs. Second, uniform sampling has been shown to benefit representation learning~\cite{shi2023how, Kang2020Decoupling}. However, in ABMIL's attention mechanism, attention is often highly skewed (See Fig. \ref{fig:attn_dist}), making it difficult to achieve uniform sampling. It consistently identifies and overweights instances deemed informative. As a result, most patches receive negligible attention weights and contribute minimally to the patch representation learning process. In contrast, our feature induced stream promotes uniform learning across all patches, potentially yielding higher-quality representations.

\subsection{Model Architecture}

    To effectively exploit spatial relationships for regularization, we need a model that can capture structural dependencies among WSI patches while efficiently processing their sparse and irregular distributions resulting from background filtering that removes uninformative non-tissue regions. Graph Attention Networks (GATs) naturally meet these requirements through their ability to operate on graph-structured data while maintaining computational efficiency. We, therefore, design our framework based on GATs, incorporating both label-guided knowledge learning and feature-induced knowledge learning streams, where the latter serves as regularization. 

    A WSI is first decomposed into N patches, each encoded into a feature vector \( \mathbf{x}_i \in \mathbb{R}^D \) using a frozen patch feature extractor. The feature set \( X = [\mathbf{x}_1, \mathbf{x}_2, \ldots, \mathbf{x}_N] \) constitutes the node set \( \mathcal{V} = \{ v_1, v_2, \ldots, v_N \} \) of the graph \( \mathcal{G} \). Edges \( \mathcal{E} \) are established based on spatial proximity: for each node \( v_i \in \mathcal{V} \), its neighborhood \( \mathcal{N}(v_i) \) includes nodes \( v_j \) if \( \text{dist}(v_i, v_j) \leq 2\sqrt{2} \), connecting nodes within a $5\times5$ grid to balance context and cost. Consequently, \( \mathcal{E} \) consists of directed edges \( e_{ji} \) from \( v_j \) to \( v_i \) for all \( v_j \in \mathcal{N}(v_i) \). This graph construction captures the topological relationships among patches. The model consists of three modules: a \text{shared} encoder that learns rich structural representations, a decoder that reconstructs patch features for self-supervised learning, and a classifier that predicts slide-level labels.

    \subsubsection{Encoder Module and Process}

    Building upon our motivation to leverage spatial relationships, the encoder employs GAT layers to capture both local and global contextual information from WSIs. Unlike traditional MIL approaches that treat patches independently, our GAT-based encoder explicitly models patch interactions through attention mechanisms. For the $(l+1)^{\text{th}}$ layer in a series of $L$ layers, the feature vector of node $i$, $H^{(l+1)}_i$, is updated by aggregating features from its neighborhood $\mathcal{N}(i)$:
    
    \begin{equation}
    H^{(l+1)}_i = \sigma\left(\bigoplus_{m=1}^{\text{num\_heads}} \sum_{j \in \mathcal{N}(i) \cup \{ i \}} \alpha_{ijm}^{(l)} \mathbf{W}_m^{(l)} H^{(l)}_j\right), \label{eq:multihead_GAT_update}
    \end{equation}
    where $\sigma$ is the activation function, $\bigoplus$ denotes concatenation across all attention heads. The attention coefficients $\alpha_{ijm}^{(l)}$ for the $m^{\text{th}}$ head are computed as:
    
    \begin{equation}
    \alpha_{ijm}^{(l)} = \frac{e^{\sigma\left(\mathbf{a}^{(l)T} \left[\mathbf{W}_m^{(l)} H^{(l)}_i \| \mathbf{W}_m^{(l)} H^{(l)}_j\right]\right)}}{\sum_{k \in \mathcal{N}(i) \cup \{ i \}} e^{\sigma\left(\mathbf{a}^{(l)T} \left[\mathbf{W}_m^{(l)} H^{(l)}_i \| \mathbf{W}_m^{(l)} H^{(l)}_k\right]\right)}} ,
    \label{eq:attention_coeff}
    \end{equation}
    where $\mathbf{W}_m^{(l)}$ is the learnable weight matrix for the $m$-th attention head in the $l$-th layer, $\mathbf{a}^{(l)}$ is a learnable weight vector that parameterizes the attention mechanism, and $\|$ denotes the concatenation operation.

    Residual connections~\cite{he2016deep} and layer normalization~\cite{ba2016layer} are applied at each layer to stabilize the training:
    
    \begin{equation}
    H^{(l+1)}_i = \text{LayerNorm}\left(H^{(l+1)}_i + H^{(l)}_i\right)
    \label{eq:layer_norm_residual}
    \end{equation}

    \subsubsection{Decoder Module and Process}
    
    To implement our proposed label-independent regularization strategy, the decoder module performs feature reconstruction through a mirrored GAT architecture while maintaining feature space alignment by omitting layer normalization in the final layer. Given masked input graph $G_m$, features are updated as:
    
    \begin{equation}
    \hat{H}^{(l+1)}_i = \sigma\left(\bigoplus_{m=1}^{\text{num\_heads}} \sum_{j \in \mathcal{N}(i) \cup \{ i \}} \beta_{ijm}^{(l)} \mathbf{V}_m^{(l)} \hat{H}^{(l)}_j\right), \label{eq:decoder_GAT_update}
    \end{equation}
    where $\beta_{ijm}^{(l)}$ is computed analogously to Eq.\ref{eq:attention_coeff} using the decoder's own set of parameters, and $\mathbf{V}_m^{(l)}$ represents the learnable weight matrix for the $m$-th attention head. This reconstruction process leverages inherent spatial patterns in WSIs as a label-independent regularizer, constraining the model's parameter space.
    
    \subsubsection{Classifier Module and Process}

    The classifier module combines global contextual information through a parameterized global node that connects to all patches in $\mathcal{G}$. This node serves as an attention-based pooling mechanism, adaptively aggregating information from diagnostically relevant regions:
    
    \begin{equation}
    H^{(L)}_{global} = \text{Encoder}(\mathcal{G} \oplus v_{global})[global],
    \end{equation}
    where $H^{(L)}_{global}$ embodies the $\mathcal{G}$'s hidden state. $\oplus$ denotes the operation of concatenating the global node $v_{global}$ to $\mathcal{G}$. The final classification is obtained through:
    
    \begin{equation}
    p_{class} = \text{MLP}(H^{(L)}_{global}).
    \end{equation}

\subsection{Learning Strategy and Target}
    \subsubsection{Dual-source Supervision}
    Our framework leverages supervision signals from two complementary sources. The first source comes from slide-level annotations, providing explicit discriminative signals for classification. The second source stems from masked patch feature prediction, where each masked patch position creates a uniformly distributed prediction target. Specifically, we randomly mask 70\% of nodes following a hypergeometric distribution, aligned with established practices in spatial reconstruction tasks~\cite{he2022masked,xie2022simmim}. The ablation on mask ratios is provided in Section~\ref{sec:moreab} of the Supplementary Material. As each position has an equal probability of being masked and subsequently predicted, this process generates diverse and uniform supervision signals across all patches in the WSI. Prior to encoding, all node features undergo a linear transformation to promote feature diversity~\cite{han2022vision}, addressing potential domain gaps from fixed feature extractors.

    \subsubsection{Label-Guided Knowledge Learning Stream}
    The label-guided stream aims to learn discriminative representations from slide-level annotations. The encoder-derived latent representations are exploited to discern discriminative patterns under label-driven supervision. This mechanism involves processing the entirety of the graph to infuse the classification process with explicit label information, thereby refining the global representation toward achieving maximal class discriminability:
    \begin{equation}
    \mathcal{L}_{comp} = -\sum_{c=1}^{C} y_{c} \log(p_{c}),
    \end{equation}
    where $y_{c}$ represents the ground truth label for $\mathcal{G}$, and $p_{c}$ denotes the classifier's output probability that $\mathcal{G}$ belongs to class $c$. This stream provides essential supervision for the downstream classification task through clean gradient flow from labels. However, the supervision signals are limited by the fixed slide-level annotations. 

    \subsubsection{Feature-Induced Knowledge Learning}
    This stream utilizes uniformly distributed supervision signals through random masked feature reconstruction. Specifically, 70\% of the node features are randomly masked in each input graph to encourage the model to learn robust representations. The encoder first processes the masked graph $\mathcal{G}_m$ to generate latent representations capturing spatial information. These latent representations then flow through two parallel paths for reconstruction and classification.

    The decoder receives the encoded $\mathcal{G}m$ and reconstructs the original graph $\mathcal{G}$. The reconstruction loss is computed at masked node positions using cosine distance:
    
    \begin{equation}
    \hat{v}_i = Decoder(Encoder(\mathcal{G}_m))[i],
    \end{equation}
    

    \begin{equation}
    \mathcal{L}_{recon}
    = \frac{1}{\lvert N_{\text{masked}} \rvert}
      \sum_{i \in N_{\text{masked}}}
      \left( 1 - \frac{ v_i \cdot \hat{v}_i }{ \lVert v_i \rVert_2 \, \lVert \hat{v}_i \rVert_2 } \right),
    \end{equation}
    where $v_i$ represents the original feature vector of the masked node $i$, $\hat{v}_i$ is its reconstructed feature vector, $N_{masked}$ is the set of \text{indices of the} masked nodes, \text{and $\lVert \cdot \rVert_2$ denotes the $\ell_2$ norm}. We adopt cosine distance for $\mathcal{L}_{recon}$ instead of MSE due to its scale-invariance property. Patch features extracted by different pretrained networks may have varying norms. Cosine distance, being invariant to feature magnitude, allows the model to focus on learning meaningful spatial relationships rather than being affected by the scale differences in feature representations. This choice makes our method more robust and adaptable across different feature extractors. Functionally, this loss serves as a
    label-agnostic regularizer that constrains the latent space itself, unlike classic regularizers that penalize weights' norms or drop activations. This prevents overfitting to sparse label-driven signals by forcing the model to learn intrinsic spatial patterns, providing a source of noise-free regularization for supervised learning.
    
    In parallel, the classifier processes these encoded representations, introducing an auxiliary prediction task on $\mathcal{G}_m$. The loss function for this task is defined as:
    
    \begin{equation}
    \mathcal{L}_{corr} = -\sum_{c=1}^{C} y_{c} \log(\hat{p}_{c}),
    \end{equation}
    where $\hat{p}_{c}$ denotes the probability that $\mathcal{G}_m$ belongs to class $c$. This auxiliary supervision, inspired by recent works~\cite{zhang2022dtfd, Tang_2023_ICCV}, improves model robustness by exposing it to diverse partial views of WSIs. More importantly, by sharing the same encoded representations between reconstruction and classification tasks, it bridges these two objectives and guides the reconstruction to maintain class-discriminative information, providing a shared optimization direction that ensures the reconstructed features not only preserve spatial patterns but also encode diagnostically relevant information critical for classification.

\subsubsection{Joint Objective Function}
The joint objective function combines the self-supervised reconstruction loss $\mathcal{L}_{recon}$, corrupted graph prediction  loss $\mathcal{L}_{corr}$, and raw supervised classification loss $\mathcal{L}_{comp}$ with weighted summation:

\begin{equation}
\mathcal{L} = \lambda_{recon} \mathcal{L}_{recon} + \lambda_{comp} \mathcal{L}_{comp} + \lambda_{corr} \mathcal{L}_{corr},
\end{equation}
where $\lambda_{recon}$, $\lambda_{comp}$, and $\lambda_{corr}$ balance the contributions of reconstruction, complete graph classification, and corrupted graph prediction losses, respectively.

\begin{table*}[h]
\caption{Comparison of methods across datasets using different feature extractors. Values are means over 5 runs; best in \textbf{bold}, second best \uline{underlined}. For BRACS dataset, AUC refers to Macro AUC.}
\label{tab:results_combined}
\centering
\begin{tabular}{lcccccccccccc}
\toprule
& \multicolumn{6}{c}{\textbf{ResNet as Feature Extractor}} & \multicolumn{6}{c}{\textbf{ViT as Feature Extractor}} \\
\cmidrule(r){2-7} \cmidrule(r){8-13}
Method & \multicolumn{2}{c}{CAMELYON-16} & \multicolumn{2}{c}{TCGA-Lung} & \multicolumn{2}{c}{BRACS} & \multicolumn{2}{c}{CAMELYON-16} & \multicolumn{2}{c}{TCGA-Lung} & \multicolumn{2}{c}{BRACS} \\
\cmidrule(r){2-3} \cmidrule(r){4-5} \cmidrule(r){6-7} \cmidrule(r){8-9} \cmidrule(r){10-11} \cmidrule(r){12-13}
& Acc & AUC & Acc & AUC & Acc & AUC & Acc & AUC & Acc & AUC & Acc & AUC \\
\midrule
\multicolumn{13}{l}{\textbf{ABMIL based}} \\
ABMIL & 0.867 & 0.880 & 0.843 & 0.928 & 0.639 & 0.790 & 0.879 & 0.887 & 0.877 & 0.944 & 0.586 & 0.796 \\
CLAM-SB & 0.881 & 0.875 & 0.857 & 0.925 & 0.616 & 0.770 & 0.874 & 0.874 & 0.874 & 0.952 & 0.614 & 0.798 \\
CLAM-MB & 0.870 & 0.884 & 0.842 & 0.935 & 0.623 & 0.794 & 0.887 & 0.906 & 0.879 & 0.952 & 0.605 & 0.783 \\
DSMIL & 0.871 & 0.874 & 0.837 & 0.936 & 0.602 & 0.785 & 0.874 & 0.886 & 0.887 & 0.959 & 0.623 & 0.795 \\
DTFD-MIL & 0.843 & \uline{0.892} & \uline{0.872} & 0.937 & 0.614 & 0.788 & 0.879 & \uline{0.911} & 0.880 & 0.950 & 0.593 & 0.788 \\
MHIM-MIL & 0.876 & 0.871 & 0.855 & 0.939 & 0.634 & 0.773 & 0.876 & 0.903 & \uline{0.894} & 0.956 & 0.577 & 0.795 \\
ACMIL & 0.876 & 0.886 & 0.845 & 0.916 & 0.641 & 0.782 & \uline{0.893} & 0.886 & 0.881 & 0.948 & 0.611 & 0.810 \\
\midrule
\multicolumn{13}{l}{\textbf{Transformer / Mamba based}} \\
TransMIL & 0.868 & 0.883 & 0.863 & 0.927 & 0.634 & 0.802 & 0.868 & 0.892 & 0.882 & 0.950 & 0.600 & 0.773 \\
RRT & 0.884 & 0.889 & 0.858 & \uline{0.941} & 0.646 & 0.808 & 0.873 & 0.895 & 0.885 & 0.952 & 0.595 & 0.787 \\
MambaMIL & \uline{0.885} & 0.888 & 0.860 & 0.932 & \uline{0.671} & 0.821 & 0.891 & 0.909 & 0.890 & \uline{0.960} & 0.634 & 0.815 \\
\midrule
\multicolumn{13}{l}{\textbf{GNN based}} \\
PatchGCN & 0.860 & 0.837 & 0.815 & 0.906 & 0.667 & \textbf{0.837} & 0.879 & 0.904 & 0.879 & 0.956 & \uline{0.651} & \uline{0.819} \\
SRMIL (Ours) & \textbf{0.912} & \textbf{0.913} & \textbf{0.878} & \textbf{0.945} & \textbf{0.676} & \uline{0.828} & \textbf{0.913} & \textbf{0.918} & \textbf{0.897} & \textbf{0.962} & \textbf{0.660} & \textbf{0.839} \\
\bottomrule
\end{tabular}
\end{table*}
\section{Experiments}
This section presents the experimental evaluation of our proposed method on three publicly available datasets, each corresponding to a distinct classification task: CAMELYON-16~\cite{bejnordi2017diagnostic} for binary tumor detection, TCGA-Lung for tumor subtyping, and BRACS~\cite{brancati2022bracs} for tissue grading ranging from normal to atypical to tumor. We adhered to official data splits when available to
ensure proper patient-level separation among splits, preventing potential information leakage from having slides from the same patient across different splits, which could lead to overly optimistic performance estimates. To further ensure an unbiased evaluation, we do not use additional threshold search or cross-validation to report the best average performance on the evaluation sets. This approach avoids practices that may inflate performance metrics, thereby providing a more accurate estimate of the model's generalization ability in real-world applications. To verify the effectiveness of our approach across different feature extractors, we conducted experiments using two extractors: a CNN-based ResNet50~\cite{he2016deep} pretrained on ImageNet and a ViT~\cite{huang2023visual} trained with self-supervision.

\subsection{Results}
We evaluate and compare our method with state-of-the-art approaches on the three datasets, running each method five times per dataset. We report the mean of the evaluation metrics, accompanied by the minimum and maximum values in parentheses. Table~\ref{tab:results_combined} presents the results.

In traditional positive vs. negative MIL tasks, the performance of DTFD-MIL is inferior to that of the ABMIL under the same settings. This suggests that the additional sub-bag loss function introduced by DTFD-MIL, which divides a bag into multiple sub-bags and assumes that the labels of the sub-bags are identical to the original bag, may lead to performance degradation. This degradation occurs as incorrect supervisory signals are easily introduced when the proportion of positive instances in positive bags is low, causing many sub-bags with only negative instances to be mistakenly trained as positive. However, DTFD-MIL exhibits strong performance in cancer subtyping tasks, where there are no distinctly positive or negative bags but only tumors of different categories. In this scenario, the additional loss introduced by DTFD-MIL is unlikely to impact the training negatively, as it employs nearly correct low-noise supervisory signals that serve as effective regularization terms. It is worth noting that on the TCGA-Lung subtyping, all methods exhibit higher AUC scores than their corresponding accuracy values. This suggests that the default threshold of 0.5 may not be optimal for this dataset and that threshold search could potentially yield higher accuracy scores. Our method shows competitive performance across all tasks and different types of feature extractors, highlighting its versatility and robustness.

\subsection{Evaluating Feature Transformation Ability}

To evaluate the quality of the embedding space, we conducted a instance classification experiment using the Camelyon16 dataset. We compared original ResNet50 features with intermediate features from ABMIL and our SRMIL. We utilized a training-free, top-5 K-Nearest Neighbor classifier. As shown in Table~\ref{tab:patch_classification}, all three feature sets achieve high accuracy scores due to the highly imbalanced nature of the instance classification task in the Camelyon16 dataset. However, the SRMIL-transformed features demonstrate notably higher recall and F1 scores compared to the other two feature sets. The higher recall suggests that our SRMIL method is more effective in reducing the instance-level false negative rate, which is of critical importance in clinical settings where missing a positive case can have severe consequences. As observed in Figure~\ref{fig:attn_dist}, the highly skewed distribution of maximum attention weights per bag during training shows that model updates in ABMIL are commonly influenced by a few instances. This leads to updates that may not reflect the full context of the data. These empirical results align with our motivation that while uniform sampling benefits representation learning, ABMIL's highly skewed attention distribution makes this difficult to achieve. Our feature-induced learning stream naturally promotes uniform learning through label-independent regularization, yielding representations that better capture the underlying tissue structure and also enable more comprehensive learning from the entire bag rather than focusing only on dominant instances.

\begin{table}
  \centering
  \caption{Metrics for instance classification using a KNN classifier}
  \label{tab:patch_classification}
  \begin{tabular}{lccc}
    \toprule
    Representation & Accuracy & Recall & F1 Score\\
    \midrule
    Original & 0.972 & 0.393 & 0.555 \\
    After ABMIL & \uline{0.973} & \uline{0.432} & \uline{0.593} \\
    After SRMIL & \textbf{0.979} & \textbf{0.569} & \textbf{0.707} \\
    \bottomrule
  \end{tabular}
\end{table}

\section{Ablation Study}

    To elucidate the effectiveness of the feature-induced learning stream, we conduct an ablation study to examine the impact of two loss functions on overall model performance. We use the CAMELYON-16 dataset in our experimental setup and employ ResNet50 as the feature extractor. Table~\ref{tab:ablation_study_0} summarizes our findings. With both losses disabled, our model reduces to a standard GAT with label-guided stream only, achieving 86.5\% accuracy. Despite utilizing spatial topology through graph structure, this performance is similar to traditional MIL methods, indicating incomplete exploitation of positional information.

    \subsection{Individual Impact}

    The corrupted graph prediction mechanism acts as data augmentation similar to dropout or sub-bagging, improving accuracy by 0.9\%. While this task relies on limited supervisory signals from label information—potentially introducing noise if all positive instances are masked—the benefits of comprehensive label information utilization outweigh these drawbacks.
    Feature reconstruction provides dual benefits in our framework. First, it encourages a well-structured feature space that facilitates learning. Second, it acts as a regularization mechanism against overfitting caused by limited label information. This task yields a more substantial improvement of 3.1\%, demonstrating its effectiveness in further exploiting positional information.

    \subsection{Synergistic Effect}
    \label{subsec:synergy}

    The integration of these two complementary tasks leads to a notable performance gain beyond their individual contributions, achieving an accuracy of 91.2\%. This improvement suggests a unified optimization direction towards more discriminative feature learning. The reconstruction process benefits from the corrupted prediction task by encoding features that decode into more discriminative information rather than simply preserving average structural information. Moreover, the improvement indicates that the gradient flows from these two objectives exhibit minimal interference, allowing both tasks to optimize effectively in a joint feature space. This synergy highlights how feature-induced stream enables more complete exploitation of WSI information. We also perform an ablation without the classification loss on uncorrupted images. Surprisingly, the performance does not degrade substantially. One possible explanation is that, although the classifier never observes a complete graph during training, the knowledge about the complete graph is implicitly captured through the reconstruction task. Since reconstruction progressively makes the corrupted graph resemble the complete one, the classifier trained only on corrupted graphs can still achieve robust performance. This observation further validates the synergistic effect.

\begin{table}
  \centering
  \caption{Ablation study results showcasing the impact of graph reconstruction and corrupted graph prediction components on model performance.}
  \vspace{-0.4cm}
  \label{tab:ablation_study_0}
  \resizebox{\linewidth}{!}{
  \begin{tabular}{@{}ccccc@{}}
    \toprule
    \text{$\mathcal{L}_{comp}$} & \text{$\mathcal{L}_{recon}$} & \text{$\mathcal{L}_{corr}$} & \text{Accuracy} &  \text{AUC} \\
    \midrule
    \cmark & \xmark & \xmark & 0.865 {\scriptsize(0.845,0.899)} & 0.832 {\scriptsize(0.760,0.897)} \\
    \cmark & \xmark & \cmark & 0.874 {\scriptsize(0.860,0.884)} & 0.856 {\scriptsize(0.821,0.886)} \\
    \cmark & \cmark & \xmark & \uline{0.896} {\scriptsize(0.876,0.922)} & 0.900 {\scriptsize(0.875,0.920)} \\
    \xmark & \cmark & \cmark & 0.888 {\scriptsize(0.860,0.907)} & \uline{0.903} {\scriptsize(0.866,0.926)} \\
    \cmark & \cmark & \cmark & \textbf{0.912} {\scriptsize(0.884,0.938)} & \textbf{0.913} {\scriptsize(0.889,0.929)} \\
    \bottomrule
  \end{tabular}
  }
\end{table}

\section{Conclusion}

In this paper, we demonstrate that spatial patterns inherent in whole slide images can serve as powerful label-independent regularization signals for weakly supervised learning. Through a dual-path learning architecture that jointly optimizes spatial reconstruction and label-guided classification objectives, our framework effectively addresses the overfitting challenges in computational pathology without relying on potentially noisy label-driven regularization. Our results not only validate the effectiveness of spatial patterns as regularization signals but also establish a promising paradigm for incorporating other forms of self-supervised and label-independent information into WSI analysis. Future research directions include exploring more sophisticated masking strategies, extending our approach to multi-scale and multi-modal information, and investigating more advanced graph neural network architectures and SSL techniques to improve performance and scalability further.

\section*{Acknowledgment}
This study is supported by the Department of Defense grant HT9425-23-1-0267.

{
    \small
    \bibliographystyle{ieeenat_fullname}
    \bibliography{main}
}
\clearpage
\setcounter{page}{1}
\maketitlesupplementary

\section{Attention Issues}

To better understand this issue and its implications, we conducted a detailed analysis on the Camelyon16 dataset, which provides pixel-level annotations that allow us to obtain patch-level labels for evaluation. The attention values in the ABMIL method tend to be highly skewed, as discussed in the main text and its Figure~\ref{fig:training_metrics}. Often, only a few instances receive significant attention, while most instances are neglected. Many existing WSI visualization techniques, such as the one adopted in CLAM~\cite{lu2021data}, apply a percentage transformation formula when visualizing attention scores:
\begin{equation*}
P(s_i) = \frac{R(s_i)}{n} \times 100
\end{equation*}
where $P(s_i)$ is the percentile for score $s_i$, $R(s_i)$ represents the rank of score $s_i$. This ranking method artificially transforms the distribution of attention weights to be uniform (with equal intervals). While this transformation allows for the creation of some ideal visualizations, it alters the original attention distribution in a way that cannot be reversed and obscures the real attention distribution issues. To better understand the attention distribution and its implications, instead of visualizing transformed attention weights, we visualize the raw attention weights of a negative example in Figure~\ref{fig:example}. The instance with the highest attention weight (red) accounts for 91.3\% of the total attention, while the remaining 1738 instances (blue) share the remaining 8.7\%.

\begin{figure}[!h]
\centering
\includegraphics[width=0.5\textwidth]{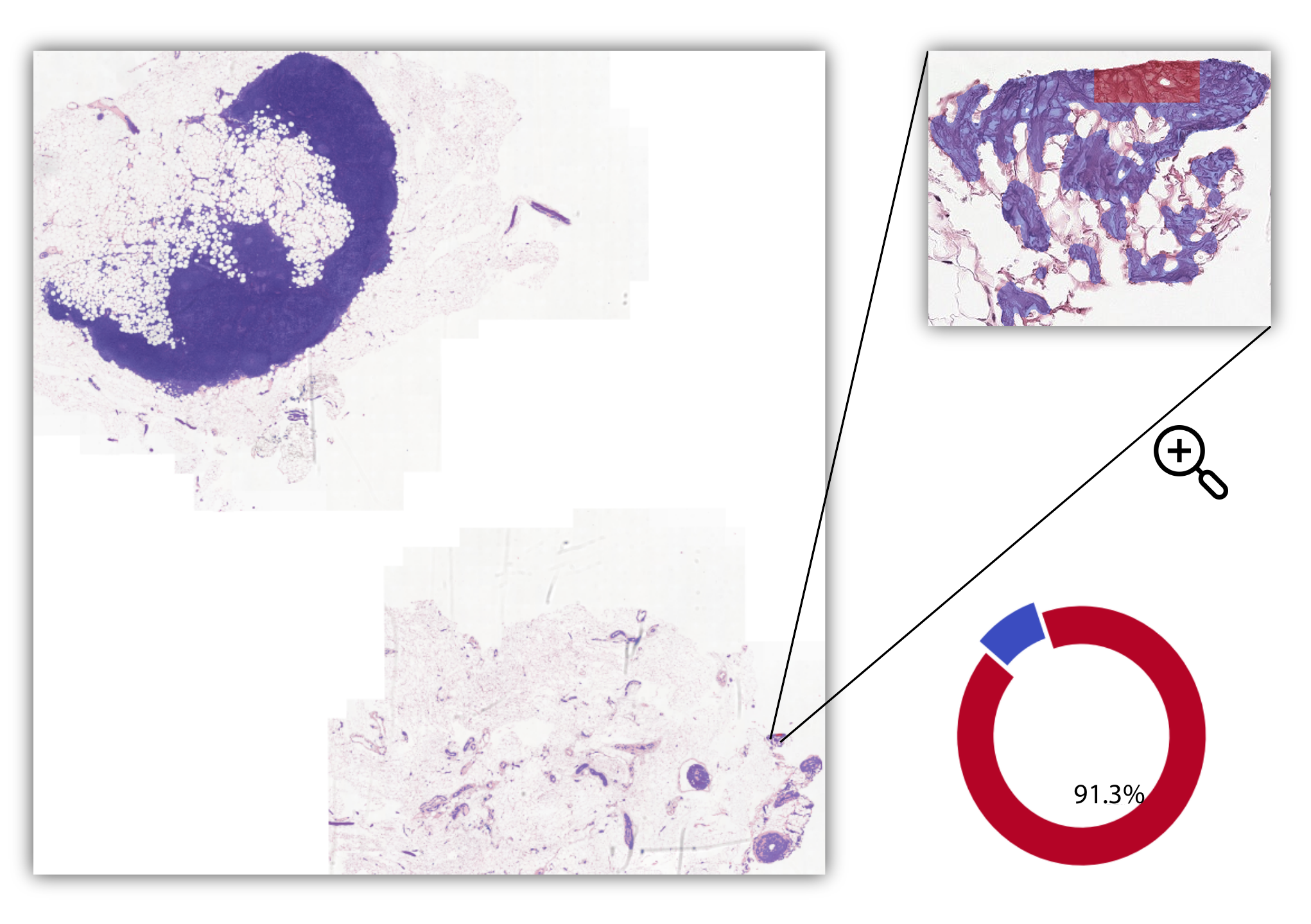}
\caption{Visualization of ABMIL attention for a negative bag.}
\label{fig:example}
\end{figure}

Ideally, in negative cases, all patches should be learned as negative and have uniform weights to ensure that information from each patch is reasonably considered during gradient updates. However, the highly skewed attention weights in ABMIL allow only a small number of instance features in each bag to be used for updating the model. The model remembers the bag by memorizing the features of only a few instances, leading to underutilization of information in negative instances. This increases the possibility of the model overfitting negative instances into positive instances to remember positive bags.

\section{Datasets}

We evaluate our proposed method on three diverse and challenging histopathology WSI datasets, each representing a unique classification task within the domain:

\begin{itemize}

    \item \textbf{CAMELYON-16 \cite{bejnordi2017diagnostic}:} This dataset is utilized for binary classification to detect tumor presence within lymph node sections. It consists of 270 training slides and 129 testing slides, with a total of 4,647,505 patches. The task assesses the model's proficiency in differentiating normal and tumor tissues. As CAMELYON-16 lacks a predefined validation set, we allocate 15\% of the training set for validation purposes.
    
    \item \textbf{The Cancer Genome Atlas (TCGA) Lung Cancer:} This dataset aims at classifying Lung Squamous Cell Carcinoma (LUSC) versus Lung Adenocarcinoma (LUAD). It contains 1,044 slides, from which 17,407,256 patches are extracted. This dataset evaluates the model's ability to distinguish between related cancer subtypes. In the absence of a predefined train-test split, we designate two-thirds of the samples for training, reserving 15\% for validation and the remaining one-third for testing.
    
    \item \textbf{BRACS \cite{brancati2022bracs}:} This dataset focuses on a three-class classification challenge involving normal tissue, atypical ductal hyperplasia, and malignant tumors, with 6,297,595 patches derived from 395 training, 65 validation, and 87 testing slides. It tests the model's capability in classifying tissue pathologies along a continuum of severity from benign to potentially malignant to unequivocally cancerous.
    
\end{itemize}

\section{Implementation Details}

Our training protocol utilized the AdamW optimizer with an initial learning rate of $1 \times 10^{-4}$ under cosine annealing scheduling for 100 epochs with weight decay set to $1 \times 10^{-5}$. The best checkpoint on the validation set was used for test set evaluation. 

We carefully considered dataset splitting strategies to ensure rigorous evaluation and prevent potential information leakage. For the three datasets used in our study:

\begin{itemize}
    \item \textbf{TCGA-Lung}: Without official splits, we randomly divided the dataset with a seed value of 42: one-third for testing and 15\% of the remaining data for validation. 
    
    \item \textbf{BRACS}: We utilized the official training, validation, and test splits provided by the dataset authors, which maintain proper patient-level separation.
    
    \item \textbf{Camelyon16}: We used the official training and test splits, and randomly allocated 15\% of the training set for validation using a seed value of 42.
\end{itemize}

When using official splits (BRACS and Camelyon16), we benefit from proper patient-level separation, where all WSIs from the same patient are confined to a single split. This separation is crucial for realistic performance evaluation, as mixing WSIs from the same patient across different splits could lead to patient-level information leakage and potentially overestimate the model's performance in real clinical applications. For TCGA-Lung where official splits were not available, we maintained consistency in our splitting approach using fixed random seeds to ensure reproducibility. For all experiments, we report test set performance using the model checkpoint that achieved the best performance on the validation set.

For preprocessing, we adapted CLAM's pipeline~\cite{lu2021data} with several modifications. Non-overlapping patches of size 224$\times$224 were extracted, departing from CLAM's original 256$\times$256 patch size. We introduced an additional grid alignment step where patch boundaries were extended to the nearest multiple of the patch size, ensuring consistent mapping between physical coordinates and the model's input grid. This alignment eliminates potential rounding errors in patch coordinate mapping, though it may result in a slight increase in the total number of patches compared to the original CLAM preprocessing.

\begin{algorithm}[t]
    \caption{Expand Contours to Fixed Step-size Grid}
    \label{algo:grid-alignment}
    \SetKwProg{Fn}{def}{:}{\KwRet}
    \KwSty{global} step\_size = 224 \\
    \Fn{extend\_contour(start\_x, start\_y, w, h)}{
        w += start\_x \% step\_size \\
        h += start\_y \% step\_size \\
        start\_x -= start\_x \% step\_size \\
        start\_y -= start\_y \% step\_size \\
        \KwSty{return} start\_x, start\_y, w, h 
    }
    \comment{contour: (start\_x, start\_y, w, h) }\\
    contour = extend\_contour(contour)
\end{algorithm}

For the reconstruction task, we use cosine distance as the loss function ($\mathcal{L}_{recon}$ in Algorithm \ref{algo:training_reconstruction}). Unlike pixel reconstruction tasks, which commonly use Mean Squared Error (MSE) as the loss function due to the fixed distribution and value range of input features, our feature reconstruction task involves input scales that might vary significantly among different feature extractors. Therefore, using cosine distance provides a more scale-invariant objective compared to MSE loss.

The models were trained using an AdamW optimizer with a learning rate of 1e-4 and a weight decay of 1e-5 for 100 epochs, coupled with cosine annealing for learning rate scheduling. The hyperparameters for the loss function were set as follows: $\lambda_{recon} = 1.8$, $\lambda_{corr} = 0.1$, and $\lambda_{comp} = 0.1$. This choice was guided by the goal of approximately balancing the gradient magnitudes between losses at the start of training, so that neither dominates the optimization. This joint optimization allows the feature-induced stream to regularize the encoder and support the formation of a robust latent space, while maintaining discriminative capability through label guidance. We used these weights across all experiments.

\begin{algorithm}[t]
    \caption{Simplified Training \& Evaluation Process}
    \label{algo:training_reconstruction}
    \comment{Training} \\
    model.train() \\
    \For{$\mathcal{G}$ in data\_loader}{
        ~\\
        $\mathcal{G}_m$ = $\mathcal{G}$ \\
        $\mathcal{G}_m$[mask\_idx] = mask\_token \\
        corr\_latent = model.encoder($\mathcal{G}_m$) \\
        \comment{reconstruction task on the corrupted view} \\
        $\mathcal{G}_{recon}$ = model.decoder(corr\_latent) \\
        $\mathcal{L}_{recon}$ = cosine\_dist($\mathcal{G}$[mask\_idx], $\mathcal{G}_{recon}$[mask\_idx]) \\
        \comment{classification task on the corrupted view} \\
        $\hat{y}_{corr}$ = model.classifier(corr\_latent) \\
        $\mathcal{L}_{corr}$ = cross\_entropy($y$, $\hat{y}_{corr}$) \\
        \comment{classification task on the complete view} \\
        $\hat{y}_{comp}$ = model.classifier(model.encoder($\mathcal{G}$)) \\
        $\mathcal{L}_{comp}$ = cross\_entropy($y$, $\hat{y}_{comp}$) \\
        loss = $\lambda_{recon}$ $\mathcal{L}_{recon}$ + $\lambda_{comp}$ $\mathcal{L}_{comp}$ + $\lambda_{corr}$ $\mathcal{L}_{corr}$ \\
        optimizer.zero\_grad() \\
        loss.backward() \\
        optimizer.step()
    }
    scheduler.step() \\
    \comment{Evaluation} \\
    model.eval() \\
    \For{$\mathcal{G}$ in data\_loader}{
        ~\\
        $\hat{y}_{comp}$ = model.classifier(model.encoder($\mathcal{G}$)) \\
        $\mathcal{L}_{comp}$ = cross\_entropy($y$, $\hat{y}_{comp}$) \\
        loss = $\mathcal{L}_{comp}$
    }
\end{algorithm}

\section{Efficiency}

\begin{table}[h!]
\centering
\caption{Computational efficiency analysis. We report the average time per WSI on the Camelyon16 dataset. All experiments were conducted on a single NVIDIA A6000 GPU.}
\label{tab:efficiency}
\begin{tabular}{@{}lcc@{}}
\toprule
Method & Training (s/WSI) & Inference (s/WSI) \\ \midrule
ATTMIL     & 0.0148 & 0.0035 \\
RRTMIL     & 0.0548 & 0.0098 \\
TransMIL   & 0.1210 & 0.0198 \\
SRMIL (Ours)     & 0.0785 & 0.0232 \\
\bottomrule
\end{tabular}
\end{table}

As shown in Table~\ref{tab:efficiency}, our SRMIL framework achieves competitive computational efficiency, requiring on average 0.0785 seconds per slide for training and 0.0232 seconds for inference. While ATTMIL is faster due to its lightweight architecture, SRMIL provides a favorable trade-off between speed and accuracy.

\section{More Ablations}
\label{sec:moreab}

\begin{figure}[!h]
\centering
\includegraphics[width=0.5\textwidth]{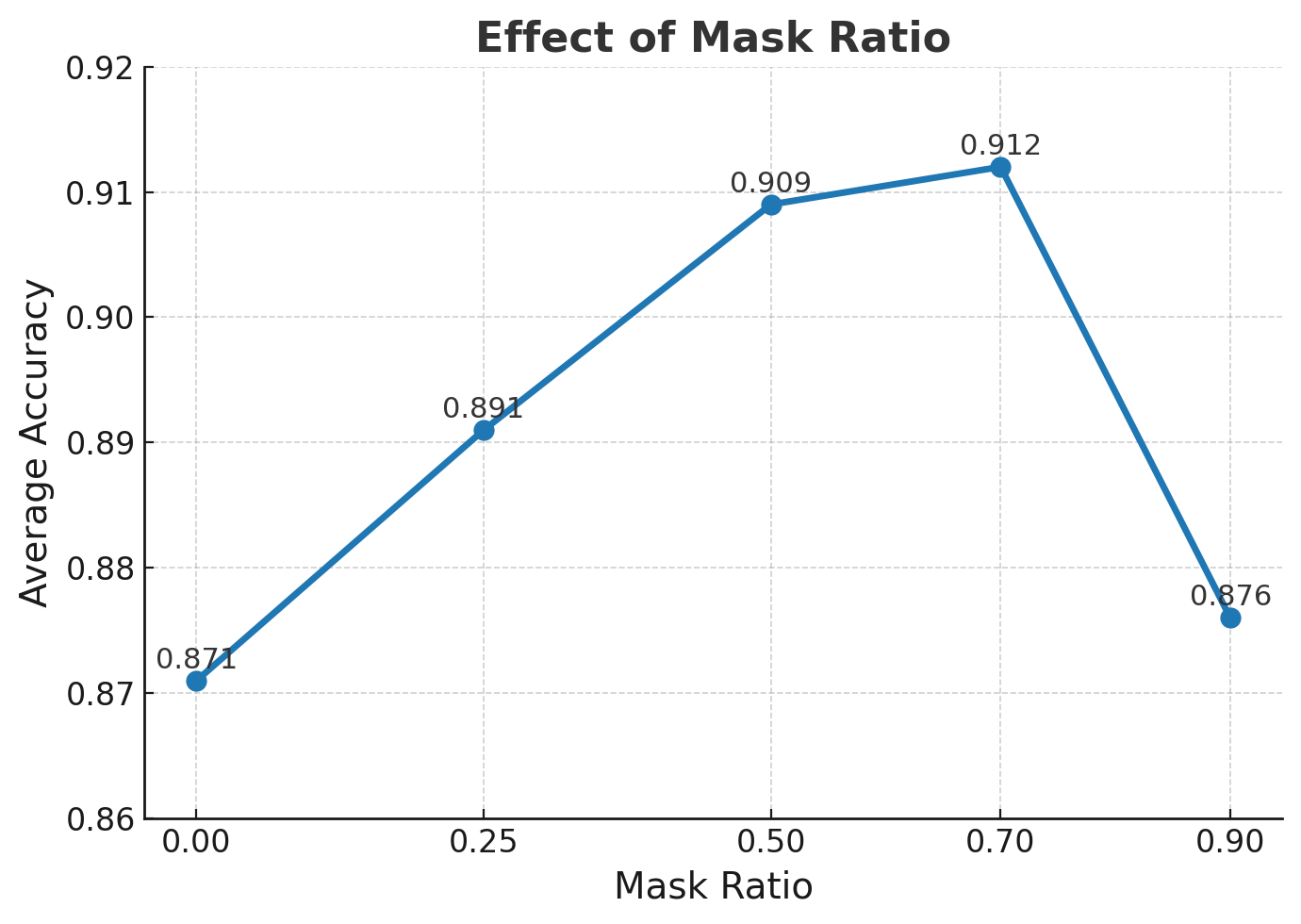}
\caption{Visualization of the effect of different mask ratios.}
\label{fig:example}
\end{figure}

When the mask ratio is set to 0, the performance (\(\sim 0.874\)) is close to that observed without applying the reconstruction loss in Tab.~\ref{tab:ablation_study_0}. Moreover, we observe that mask ratios in the range of 0.5--0.7 yield the most effective regularization, leading to consistently improved accuracy. However, further increasing the mask ratio results in performance degradation, suggesting that excessive masking cannot introduce too much useful information for latent space regularization.

\end{document}